\documentclass{article}
\usepackage{ijcai20}

\usepackage{times}
\usepackage{soul}
\usepackage{url}
\usepackage[hidelinks]{hyperref}
\usepackage[utf8]{inputenc}
\usepackage[small]{caption}
\usepackage{graphicx}
\usepackage{amsmath}
\usepackage{amsthm}
\usepackage{booktabs}
\usepackage{algorithm}
\usepackage{url}            
\usepackage{booktabs}       
\usepackage{amsfonts}       
\usepackage{nicefrac}       
\usepackage{lipsum}
\usepackage{microtype} 
\usepackage{epsfig,graphics}
\usepackage{lineno,multirow}
\usepackage{epstopdf}
\usepackage{footmisc}
\usepackage{url}
\usepackage{mathtools}

\usepackage{dsfont}
\usepackage{algorithm}
\usepackage{amsmath}
\usepackage{algpseudocode}
\usepackage{multirow}
\usepackage{graphicx}
\usepackage{threeparttable}
\usepackage{bbm}
\usepackage{tikz}           
\usepackage{xcolor}         
\usepackage{bm}
\usepackage{subcaption}
\usetikzlibrary{shapes,positioning}
\tikzset{>=latex}

\newcommand{\bz}{\mathbf{z}}
\newcommand{\bx}{\mathbf{x}}

\newcommand{\citet}[1]{\citeauthor{#1}\shortcite{#1}}
\newcommand{\citep}{\cite}

\newtheorem{prop}{Proposition}

\title{Decomposed Adversarial Learned Inference}

\author{
Alexander Hanbo Li$^{*,1}$
\and
Yaqing Wang$^{*,2}$\and
Changyou Chen$^{2}$\And
Jing Gao$^2$
\affiliations
$^1$Amazon Alexa AI\\
$^2$SUNY Buffalo
\emails
alexanderhanboli@gmail.com,
\{yaqingwa, changyou, jing\}@buffalo.edu
}

\begin{document}

\maketitle


\begin{abstract}
  Effective inference for a generative adversarial model remains an important and challenging problem. We propose a novel approach, Decomposed Adversarial Learned Inference (DALI), which explicitly matches prior and conditional distributions in both data and code spaces, and puts a direct constraint on the dependency structure of the generative model. We derive an equivalent form of the prior and conditional matching objective that can be optimized efficiently without any parametric assumption on the data. We validate the effectiveness of DALI on the MNIST, CIFAR-10, and CelebA datasets by conducting quantitative and qualitative evaluations. Results demonstrate that DALI significantly improves both reconstruction and generation as compared to other adversarial inference models.
\end{abstract}

\section{Introduction}
\label{sec:intro}
Deep directed generative models like variational autoencoder (VAE) \cite{kingma2013auto,rezende2014stochastic} and generative adversarial network (GAN) \cite{goodfellow2014generative} have been proved to be powerful for modeling complex high-dimensional distributions. While both VAE and GAN can learn to generate realistic images, their underlying mechanisms are fundamentally different. VAE maps the data into low-dimensional codes using an encoder, and then reconstructs the original data by a decoder. This allows it to perform both generation and inference. GAN, on the other hand, trains a generator and a discriminator adversarially. The generator learns to fool the discriminator by mapping low-dimensional noise vectors to the data space; at the same time, the discriminator evolves to detect the generated fake samples from the true ones. These two methods have complementary strengths and weaknesses. VAE can learn a bidirectional mapping between data and code spaces, but relies on over-simplified parametric assumptions on the complex data distribution, thereby causing it to generate blurry images \cite{donahue2016adversarial,goodfellow2014generative,larsen2015autoencoding}. GAN generates more realistic samples than VAE \cite{radford2015unsupervised,larsen2015autoencoding} because the adversarial regime allows it to learn more complex distributions. However, note that GAN only learns a unidirectional mapping for data generation, and does not allow inferring the latent codes from given samples. This is limiting because the ability of inference is very crucial for several downstream applications, such as classification, clustering, similarity search, and interpretation. Furthermore, GAN also suffers from the mode collapse problem \cite{che2016mode,salimans2016improved} -- many modes of the data distribution are not represented in the generated samples.

Therefore, one may wonder on whether we can develop a generative model that enjoys the strengths of both GAN and VAE without their inherent weaknesses. Such model should be able to generate high-quality samples as good as GAN, have an inference mechanism as effective as VAE, and also avoid the mode collapse issue. Many recent efforts have been devoted to combining VAE with adversarial discriminator(s) \cite{brock2016neural,che2016mode,larsen2015autoencoding,makhzani2015adversarial,mescheder2017adversarial}. However, VAE-GAN hybrids tend to manifest a compromise of the strengths and weaknesses of both the approaches. The main reason is that all of them retain the VAE structure, which requires an explicit metric to measure the data reconstruction and assumes over-simplified parametric data distributions. To overcome such limitations, adversarially learned inference (ALI) \cite{donahue2016adversarial,dumoulin2016adversarially} was recently proposed, wherein the discriminator is trained on the joint distribution of data and latent codes. In this way, under a perfect discriminator, one can match joint distributions of the decoder and encoder, thereby, performing inference by sampling from the encoder's conditional that also matches the decoder's posterior. In practice, however, the equilibrium of the jointly adversarial game is hard to attain as the dependency structure between data and codes is not explicitly specified. The reconstructions of ALI are thus not always faithful \cite{dumoulin2016adversarially,li2017alice} implying that its inference is not always effective. 

To overcome the aforementioned issues, in this paper, we propose a novel approach, decomposed adversarial learned inference (DALI), that integrates efficient inference to GAN and overcomes the limitations of prior approaches. The approach keeps the structure simple, involving only one generator, one encoder, and one discriminator. Furthermore, DALI's objective is directly derived from our goal of matching both prior and conditional distributions of the generator and encoder, instead of a heuristic combination with $l_k$ norm regularization. Compared to regular GANs, DALI has the ability to conduct inference, and also does not suffer from the mode collapse problem. Moreover, DALI also abandons the unrealistic parametric assumption on the conditional data distribution, and does not require any reconstruction in the data space. This is fundamentally different from VAE or VAE-GAN hybrids in which the $l_k$ norm is used to measure the data reconstruction. The usage of simple data-fitting metrics on the complex data distribution leads to worse generation performance. Different from ALI, DALI decomposes the hard problem of matching the joint distributions into two sub-tasks -- explicitly matching the priors on the latent codes and the conditionals on the data. As a consequence of more restrictive constraint, it achieves better generation and more faithful reconstruction than ALI. 
Note that GAN variations with inference mechanism usually worse generation performance as compared to regular GANs~\cite{rosca2018distribution}. To the best of our knowledge, as demonstrated in the experiments, DALI is the first framework that further improves the generating performance compared with GANs with the same architecture, while providing consistent inference on even complicated distributions.

\section{Background}
\label{sec:background}



We consider the generative model $p_{\theta^*}(\bz)p_{\theta^*}(\bx|\bz)$, where a latent variable $\bz^{(i)}$ is first generated from the prior distribution $p_{\theta^*}(\bz)$, and then the data $\bx^{(i)}$ is sampled from the conditional distribution $p_{\theta^*}(\bx|\bz)$. The parameter $\theta^*$ stands for the ground truth parameter of the underlying distribution. The prior $p_{\theta^*}(\bz)$ is always assumed to be a simple parametric distribution (e.g. $\mathcal{N}(\mathbf{0}, \mathbf{I})$), but the generative conditional $p_{\theta^*}(\bx|\bz)$ is much more complicated and not known to us. Moreover, the posterior distribution $p_{\theta^*}(\bz|\bx)$ is intractable but stands for an important inference procedure: given a data $\bx^{(i)}$, it allows us to infer its latent variable $\bz^{(i)}$.

\section{Methodology}
\label{sec:methodology}
In our method, we will model the generating process by a neural network called \textit{generator}, and the inference process by another neural network called \textit{encoder}. Consider the following two distributions of the generator and encoder, and their corresponding sampling procedures:
\begin{itemize}
    \item the \textit{generator} distribution: $p_{\theta}(\bz)p_{\theta}(\bx|\bz)$; $\bz \sim p_{\theta}(\bz)$, $\bx \sim p_{\theta}(\bx|\bz)$.
    \item the \textit{encoder} distribution: $q_{\phi}(\bx)q_{\phi}(\bz|\bx)$; $\bx \sim q_{\phi}(\bx)$, $\bz \sim q_{\phi}(\bz|\bx)$.
\end{itemize}
The generator's conditional $p_{\theta}(\bx|\bz)$ approximates the generating distribution $p_{\theta^*}(\bx|\bz)$. The encoder's conditional $q_{\phi}(\bz|\bx)$ approximates the posterior distribution $p_{\theta^*}(\bz|\bx)$, which is what we need for inference. The marginal distribution $q_{\phi}(\bx)$ stands for the empirical data distribution, and the other marginal $p_{\theta}(\bz)$ is taken to be $p_{\theta}(\bz)$, which is always a known distribution like standard Gaussian.

\subsection{Decomposition of the Joint Distribution}
The ultimate goal is to match the joint distributions, $p_{\theta}(\bx,\bz)$ and $q_{\phi}(\bx,\bz)$. If this is achieved, we are guaranteed that all marginals match and all conditionals match as well. In particular, the conditional $q_{\phi}(\bz|\bx)$ matches the posterior $p_{\theta}(\bz|\bx)$. We propose to decompose this goal into two sub-tasks -- matching the priors $p_{\theta}(\bz)$ and $q_{\phi}(\bz)$, and matching the conditionals $p_{\theta}(\bx|\bz)$ and $q_{\phi}(\bx|\bz)$. There are two advantages. Firstly, we explicitly define the dependency structure $\bz \to \bx$. Secondly, the explicit constraints on both priors and conditionals are stronger than merely one constraint on the joint distributions.


More formally, we decompose the problem of minimizing $KL(p_{\theta}(\bx, \bz), q_{\phi}(\bx, \bz))$ into matching both the prior and conditional distributions, that is, to minimize
\begin{equation}
\label{eq:KL}
\mathbb{E}_{p_{\theta}(\bz)} KL(p_{\theta}(\bx|\bz) || q_{\phi}(\bx|\bz)) + KL(p_{\theta}(\bz) || q_{\phi}(\bz)).
\end{equation}
Note that \eqref{eq:KL} is not identical to ALI's objective, but their minimums are attained at the same point. By the properties of KL-divergence, when the minimum of \eqref{eq:KL} is attained, we have $p_{\theta}(\bz) = q_{\phi}(\bz)$ and $p_{\theta}(\bx|\bz) = q_{\phi}(\bx|\bz)$ for all $\bx$ and $\bz$, and hence $p_{\theta}(\bx, \bz) = q_{\phi}(\bx, \bz)$.

\subsection{Objective Function}
\label{subsec:objective}
The objective \eqref{eq:KL} cannot be directly optimized because both $q_{\phi}(\bz)$ and $q_{\phi}(\bx|\bz)$ are impossible to sample from, as the flow in the encoder is from $\bx$ to $\bz$. However, we prove that the intractable \eqref{eq:KL} can be rephrased as the combination of a KL-divergence term and a reconstruction term, both containing only distributions that can either be sampled from or directly evaluated.

Firstly, by definition of KL-divergence, for any fixed $\bz$,
\begin{eqnarray}\label{eq:decomp_kl_1}
&KL(p_{\theta}(\bx|\bz) || q_{\phi}(\bx|\bz)) \nonumber\\
=&\mathbb{E}_{p_{\theta}(\bx|\bz)} \left[ \log p_{\theta}(\bx|\bz) - \log q_{\phi}(\bx|\bz) \right].
\end{eqnarray}
Then by Bayes' theorem, we have $\log q_{\phi}(\bx|\bz) = \log q_{\phi}(\bx) + \log q_{\phi}(\bz|\bx) - \log q_{\phi}(\bz)$. Plugging this identity into \eqref{eq:decomp_kl_1} and doing some algebra, we get
\begin{align}\label{eq:decomp_kl}
KL(p_{\theta}(\bx|\bz) || q_{\phi}(\bx)) - \mathbb{E}_{p_{\theta}(\bx|\bz)} [\log q_{\phi}(\bz|\bx)] + \log q_{\phi}(\bz).
\end{align}

Next for the second term of \eqref{eq:KL}, we also write out the definition
\begin{align*}
    KL(p_{\theta}(\bz) || q_{\phi}(\bz)) = \mathbb{E}_{p_{\theta}(\bz)} \left[ \log p_{\theta}(\bz) - \log q_{\phi}(\bz) \right].
\end{align*}
Then we have
\begin{align}
    &\mathbb{E}_{p_{\theta}(\bz)} KL(p_{\theta}(\bx|\bz) || q_{\phi}(\bx|\bz)) + KL(p_{\theta}(\bz) || q_{\phi}(\bz)) \nonumber \\
    = & \mathbb{E}_{p_{\theta}(\bz)} [\eqref{eq:decomp_kl}] + \mathbb{E}_{p_{\theta}(\bz)} \left[ \log p_{\theta}(\bz) - \log q_{\phi}(\bz) \right] \nonumber \\
    = & \mathbb{E}_{p_{\theta}(\bz)} \left[ KL(p_{\theta}(\bx|\bz) || q_{\phi}(\bx)) - \mathbb{E}_{p_{\theta}(\bx|\bz)} \log q_{\phi}(\bz|\bx) \right] + C
\end{align}
where $C = \mathbb{E}_{p_{\theta}(\bz)} \log p_{\theta}(\bz)$ is a constant because the prior $p_{\theta}(\bz)$ is a fixed parametric distribution. For example, when $\bz \sim \mathcal{N}(\mathbf{0}, \mathbf{I}_{d})$, we have $\mathbb{E}_{p_{\theta}(\bz)} \log p_{\theta}(\bz) = -d(1 + \log(2 \pi)) / 2$. Therefore, minimizing the objective \eqref{eq:KL} is now transformed to minimizing the new objective
\begin{align}\label{eq:final_decomp}
    \mathbb{E}_{p_{\theta}(\bz)} \left\{ KL(p_{\theta}(\bx|\bz) || q_{\phi}(\bx)) + \mathbb{E}_{p_{\theta}(\bx|\bz)} \left[-\log q_{\phi}(\bz|\bx)\right] \right\}.
\end{align}
Intuitively, term $(I)=\mathbb{E}_{p_{\theta}(\bz)} KL(p_{\theta}(\bx|\bz) || q_{\phi}(\bx))$ measures the difference between the generated and real samples, and term $(II)=\mathbb{E}_{p_{\theta}(\bz)} \mathbb{E}_{p_{\theta}(\bx|\bz)} \left[-\log q_{\phi}(\bz|\bx)\right]$ measures the reconstruction of the latent codes. We summarize the above procedure as a proposition.

\begin{prop} \label{prop:1}
The final objective function \eqref{eq:final_decomp}
$\mathbb{E}_{p_{\theta}(\bz)} \{ KL(p_{\theta}(\bx|\bz) || q_{\phi}(\bx)) + \mathbb{E}_{p_{\theta}(\bx|\bz)} [-\log q_{\phi}(\bz|\bx)] \}$
is minimized when $p_{\theta}(\bz) = q_{\phi}(\bz)$ and $p_{\theta}(\bx|\bz) = q_{\phi}(\bx|\bz)$ for all $\bz$, $\bx$. And hence, at the minimum, the joint distributions $p_{\theta}(\bx, \bz) = q_{\phi}(\bx, \bz)$.
\end{prop}

\subsection{Relation to the Variational Autoencoder}
The VAE \cite{kingma2013auto} method, using our notations in this paper, actually depends on the following identity:
\begin{align}
\label{eq:VAE_decomp}
&KL(q_{\phi}(\bz|\bx)||p_{\theta}(\bz|\bx)) \\
= &KL(q_{\phi}(\bz|\bx)||p_{\theta}(\bz)) - \mathbb{E}_{q_{\phi}(\bz|\bx)}\log p_{\theta}(\bx|\bz) + \log p_{\theta}(\bx). \nonumber
\end{align}
Then because of the non-negativity of KL-divergence, we have
\begin{equation*}
    \log p_{\theta}(\bx) \ge \underbrace{\mathbb{E}_{q_{\phi}(\bz|\bx)}[\log p_{\theta}(\bx|\bz)] - KL (q_{\phi}(\bz|\bx) || p_{\theta}(\bz))}_{\text{ELBO}},
\end{equation*}
and hence maximizing the log-likelihood of the observations can be transferred to maximizing the evidence lower bound (ELBO). But taking a closer look at \eqref{eq:VAE_decomp} and comparing it to \eqref{eq:decomp_kl}, we notice that \eqref{eq:VAE_decomp} is a decomposition of the KL-divergence between two conditionals, $q_{\phi}(\bz|\bx)$ and $p_{\theta}(\bz|\bx)$. Therefore, we can follow the same approach after \eqref{eq:decomp_kl} and get the following identity:
\begin{align}
\label{eq:vae_final}
& \mathbb{E}_{q_{\phi}(\bx)} KL (q_{\phi}(\bz||\bx)||p_{\theta}(\bz|\bx)) + KL(q_{\phi}(\bx)||p_{\theta}(\bx)) \nonumber \\ & - \mathbb{E}_{q_{\phi}(\bx)} \log q_{\phi}(\bx) \nonumber \\
=& \mathbb{E}_{q_{\phi}(\bx)} \left\{ KL (q_{\phi}(\bz|\bx)||p_{\theta}(\bz)) + \mathbb{E}_{q_{\phi}(\bz|\bx)} [-\log p_{\theta}(\bx|\bz)] \right\}.
\end{align}
We denote $I_{vae} = \mathbb{E}_{q_{\phi}(\bx)} [KL (q_{\phi}(\bz|\bx)||p_{\theta}(\bz))]$ and $II_{vae} = \mathbb{E}_{q_{\phi}(\bx)} \mathbb{E}_{q_{\phi}(\bz|\bx)} [-\log p_{\theta}(\bx|\bz)]$. Since the marginal $q_{\phi}(\bx)$ stands for the empirical data distribution, the right hand side of \eqref{eq:vae_final} is the empirical expectation of the negative ELBO, which is what VAE tries to minimize. We then conclude from \eqref{eq:vae_final} that VAE performs marginal distribution matching in the data space and conditional distribution matching in the latent space. This distribution matching of VAE is also observed by \citet{rosca2018distribution}.

However, the marginal distributions in the data space are very complex, and the direction $\bx \to \bz$ in the conditional distributions in the latent space is actually opposite to the generating process $\bz \to \bx$. Hence, in order to match these distributions, VAE's objective has a reconstruction term $\text{II}_{vae}$ on $\bx$, and a regularization term $\text{I}_{vae}$ on latent $\bz$. But to evaluate both terms, we need to make parametric assumptions on both conditionals $q_{\phi}(\bz|\bx)$ and $p_{\theta}(\bx|\bz)$. The assumption on $\text{I}_{vae}$ can be loosed using GANs \cite{makhzani2015adversarial}, but the assumption on $\text{II}_{vae}$ is critical and limits the performance of VAE-GAN hybrids.

Our model, DALI, instead performs marginal distribution matching in the latent space and conditional distribution matching in the data space. From \eqref{eq:final_decomp}, since the term $\text{I}$ will be replaced with an adversarial game (see Section \ref{ssec:DALI_framework}), the only assumption we need to make is on term II, that is, on the conditional $q_{\phi}(\bz|\bx)$. And our model is very flexible in its dependence on $\bz$. This assumption is much weaker than that on $p_{\theta}(\bx|\bz)$ and does not lead to the problems of VAE or VAE-GANs (e.g. blurriness).

\subsection{DALI framework}
\label{ssec:DALI_framework}

The KL-divergence part (I) can be replaced by an adversarial game using the $f$-divergence theory \cite{nowozin2016f}. The reconstruction term (II) is a log-likelihood and can be simply evaluated if we assume a parametric $q_{\phi}(\bz|\bx)$. Therefore, our framework only requires exactly one \textit{generator G}, one \textit{discriminator D}, and one \textit{encoder E}. We will now discuss how to play the adversarial game and measure the reconstruction in details.



\paragraph{Adversarial game}
Because we do not want to make any parametric assumption on the distribution $p_{\theta}(\bx|\bz)$, an adversarial game will be played to distinguish $p_{\theta}(\bx|\bz)$ from $q_{\phi}(\bx)$. By the theory of $f$-GAN \cite{nowozin2016f}, we construct an adversarial game with the value function $V(G, D)$ to be
\begin{align} \label{eq:V_GD}
\mathbb{E}_{\bx \sim q_{\phi}(\bx)} [D(\bx)] + \mathbb{E}_{\bz \sim p_{\theta}(\bz)} [-\exp_{\theta}(D(G(\bz)) - 1)].
\end{align}
Under the perfect discriminator, finding the optimal generator of \eqref{eq:V_GD} is then equivalent to minimizing the KL-divergence. The activation function for the discriminator in \eqref{eq:V_GD} is just the identity mapping instead of the sigmoid function in the original GAN. But just like in the original GAN, the generator of \eqref{eq:V_GD} also suffers from the gradient vanishing problem \cite{goodfellow2016nips}. Therefore, in our experiments, we maximize \eqref{eq:V_GD} for the discriminator $D$, but minimize $\mathbb{E}_{\bz \sim p_{\theta}(\bz)} [-D(G(\bz))]$ for the generator $G$. We call the algorithm using this value function DALI-$f$.

As shown in \citet{fedus2017many,lucic2018gans}, the equilibrium of the adversarial is hard to attain in practice, and we are not using the theoretical value function to train $G$ because of the gradient vanishing problem. Therefore, we also try using WGAN and GAN for the adversarial game in our experiments, and find out GAN provides consistently better and more stable results.


\paragraph{Reconstruction}
Because of the simplicity of the distribution of $\bz$, we make a reasonable parametric assumption on $q_{\phi}(\bz|\bx)$ so that the log-likelihood can be explicitly calculated. In this paper we will assume $\bz|\bx \sim \mathcal{N}(\mu(\bx), \sigma^2(\bx) \mathbf{I})$, and define
\begin{equation} \label{eq:loglikelihood_E}
\begin{aligned}
&L(\bz, \mu(\bx), \sigma^2(\bx)) := -\log q_{\phi}(\bz|\bx) \\
= &\frac{1}{2} \sum_{j=1}^d \left( \frac{(z_j - \mu_j(\bx))^2}{\sigma_j^2(\bx)} + \log \sigma_j^2(\bx) + \log(2 \pi) \right),
\end{aligned}
\end{equation}
where $d$ is the dimension of the latent variable $\bz$. In this case, the encoder network only needs to output two vectors, $\mu(\bx)$ and $\sigma^2(\bx)$, that is, $E(\bx) = (\mu(\bx), \sigma^2(\bx))$. Then we can compute the approximate negative posterior log-likelihood by plugging $E(\bx)$ into \eqref{eq:loglikelihood_E}.

\paragraph{Final Framework}
To summarize, our final optimization problem is
\begin{equation}\label{eq:final_loss}
    \min_{G, E} \max_{D} \Bigl\{ V(G, D) + \lambda \mathbb{E}_{p_{\theta}(\bz)} [L(\bz, E(G(\bz)))] \Bigr\}.
\end{equation}
Here, $\lambda$ is a hyper-parameter that needs to be set so that two parts of \eqref{eq:final_loss} are in the same scale. We will discuss the selection of $\lambda$ in detail in the experiment section.

\subsection{Training and Inference Procedures}
The training procedure is summarized in Algorithm \ref{alg:DALI}. Given random $\bz^{(i)} \sim p_{\theta}(\bz)$, we first generate samples $\tilde{\bx}^{(i)} \sim p_{\theta}(\bx|\bz^{(i)})$ using the generator. Then the discriminator is updated to distinguish between generated and real samples. The encoder outputs the parameters for the distribution $q_{\phi}(\bz|\bx)$, from which we calculate the log-likelihood in (II). Then the generator and encoder are updated together to minimize the reconstruction error (i.e. maximize the expected log-likelihood), while the generator has an extra goal that is to fool the discriminator. For any data $\bx^{(i)}$, its inferred latent code is set to be the conditional mean $\mu(\bx^{(i)}) = \mathbb{E}_{q_{\phi}(\bz|\bx^{(i)})} [\bz]$. Then the reconstruction of $\bx^{(i)}$ is $G(\mu(\bx^{(i)}))$. Besides the reconstruction, we can also generate more samples which are close to $\bx^{(i)}$ in the sense that they have similar latent codes. This can be done by first sampling $\bz$'s from the posterior $q_{\phi}(\bz|\bx^{(i)})$, and then map them to the data space using the generator.

\begin{algorithm}[!htb]
\small
\begin{algorithmic}
    \State $\theta_{g}, \theta_{d}, \theta_{e} \gets \text{initialize network parameters}$
    \Repeat
		\State $\bz^{(1)}, \ldots, \bz^{(n)} \sim p_{\theta}(\bz)$
		    \Comment{Draw $n$ samples from the prior}
		\State $\tilde{\bx}^{(j)} \gets G(\bz^{(j)}),
				\quad j = 1, \ldots, n$
			\Comment{Generate samples using the generator network}
		\State $(\mu(\tilde{\bx}^{(j)}), \sigma^2(\tilde{\bx}^{(j)})) \gets E(\tilde{\bx}^{(j)})$
		    \Comment{Calculate mean and variance of $q_{\phi}(\bz|\tilde{\bx}^{(j)})$}
        \State $\rho_q^{(i)} \gets D(\bx^{(i)}),
				\quad i = 1, \ldots, n$
            \Comment{Compute discriminator predictions}
        \State $\rho_p^{(j)} \gets D(\tilde{\bx}^{(j)}),
				\quad j = 1, \ldots, n$
        \State $\mathcal{L}_d \gets
            -\frac{1}{n} \sum_{i=1}^n \log(\rho_q^{(i)})
            -\frac{1}{n} \sum_{j=1}^n\ log(1 - \rho_p^{(j)})$
            \Comment{Compute discriminator loss}
        \State $\mathcal{L}_g \gets
            -\frac{1}{n} \sum_{j=1}^n \log(\rho_p^{(j)})$
            \Comment{Compute generator loss}
        \State $\mathcal{L}_e \gets
            \frac{\lambda}{n} \sum_{j=1}^n L(\bz^{(j)}, \mu(\tilde{\bx}^{(j)}), \sigma^2(\tilde{\bx}^{(j)}))$
            \Comment{Compute encoder loss}
        \State $\mathcal{L}_{rec} \gets \mathcal{L}_g + \mathcal{L}_e$
            \Comment{Compute reconstruction loss}
        \State $\theta_d \gets \theta_d - \nabla_{\theta_d} \mathcal{L}_d$
            \Comment{Gradient update on discriminator network}
        \State $(\theta_g,\theta_e) \gets (\theta_g,\theta_e) - \nabla_{(\theta_g,\theta_e)} \mathcal{L}_{rec}$
            \Comment{Gradient update on generator and encoder networks}
    \Until{convergence}
\end{algorithmic}
\caption{\label{alg:DALI} The DALI training procedure.}
\end{algorithm}

\section{Experiments}
\label{sec:experiments}
We evaluate our proposed method, DALI, for both reconstruction and generation tasks, on the data sets MNIST \cite{lecun1998gradient}, CIFAR-10 \cite{krizhevsky2009learning} and CelebA \cite{liu2015deep}. 
To show the effectiveness of DALI on mode collapse reduction, we also conduct the same 2D Gaussian mixture experiment as in \citet{dumoulin2016adversarially}. 
The architectures of our discriminator and generator are based on DCGAN \cite{radford2015unsupervised} and slightly simpler, which can be easily replaced by more advanced state-of-the-art GANs, and we use a deterministic generator throughout the experiments. Our encoder network consists of convolutional layers followed by two separated fully connected networks, which are used to predict the mean and variance of the posterior $q_{\phi}(\bz|\bx)$, respectively. The Adam optimizer \cite{kingma2014adam} is used and the learning rate decay strategy suggested by \citet{kingma2014adam} is applied. Since there are $d$ summands in \eqref{eq:loglikelihood_E}, we simply set $\lambda$ to be $1/d$ in our experiments to calculate the average distance on each dimension of $\bz$. We also observe that the discriminator shares a similar task with the encoder: both of them need to extract higher level features from raw images. Therefore, in order to reduce the number of parameters and to stabilize the training procedure, our encoder takes the intermediate hidden representation learned by the discriminator as its own input. It is worth noting that the encoder does not update the common feature extracting layers. We use the PyTorch 1.1 to implement our model. 

\subsection{Quantitative Results on Real Datasets}
\label{subsec:quant}
In this section, we use quantitative measures (MSE, Inception Score (IS), Frechet Inception Distance (FID)) to compare the inference and generation performance of DALI, GAN, ALI and ALICE. And for fair comparison, GAN is implemented to have the identical generator and discriminator with DALI. We also include a reduced version of DALI, named DALI-$l_2$, in which the conditional distribution $q_{\phi}(\bz|\bx)$ of the encoder is assumed to be a Gaussian with identity covariance matrix.  To evaluate the performance of inference, we measure it through reconstructing test images and calculating the mean squared error (MSE), which has been adopted in \citet{li2017alice}. As for generation, we calculate the inception score \cite{salimans2016improved} on $50,000$ randomly generated images. The inception scores on MNIST are evaluated by the pre-trained classifier from \citet{li2017alice}, and the inception scores on CIFAR-10 is based on the ImageNet. The quantitative results are summarized in Table \ref{tab:MNIST_CIFAR}.

\begin{table*}[!htb]
\centering
\caption{MSE~(lower is better) and Inception scores~(higher is better) on MNIST and CIFAR-10. ALI and ALICE results are from the experiments in \protect\citet{li2017alice}.}
\label{tab:MNIST_CIFAR}
\begin{tabular}{ccccccc}
\toprule
\multirow{2}{*}{Method} &\multicolumn{2}{c}{MNIST}
&&
\multicolumn{3}{c}{CIFAR-10} \\
\cline{2-4}\cline{5-7}
 &MSE & Inception Score     && MSE &Inception Score   & FID \\
\midrule
DALI &  \textbf{0.026 $\pm$ 0.018}& \textbf{9.483 $\pm$ 0.020}  & &\textbf{0.019 $\pm$ 0.009}& \textbf{6.450 $\pm$ 0.085} & \textbf{28.4} \\
DALI-$l_2$  &0.028 $\pm$ 0.018 & 9.331 $\pm$ 0.021 & &0.037 $\pm$ 0.017 & 6.324 $\pm$ 0.056 & 29.2 \\
\cline{1-7}
GAN  & - & 9.464 $\pm$ 0.020 & & -  &6.287 $\pm$ 0.061\ & 37.1 \\
ALI  & 0.480 $\pm$ 0.100 & 8.749 $\pm$ 0.090  & &0.672 $\pm$ 0.113 & 5.930 $\pm$ 0.044 & 58.9\\
ALICE  &0.080 $\pm$ 0.007& 9.279 $\pm$ 0.070 & &0.416 $\pm$ 0.202 & 6.015 $\pm$ 0.028 & -\\
\bottomrule
\end{tabular}
\end{table*}

\paragraph{Inference} From Table \ref{tab:MNIST_CIFAR}, DALI achieves the best reconstruction results on both data sets. On MNIST, DALI significantly decreases the MSE by 68\% and 95\% compared with ALICE and ALI respectively. On the more complicated CIFAR-10 data set, DALI decreases the MSE by 95\% and 97\%. In order to alleviate the non-identifiable issue of ALI, ALICE adds the conditional entropy constraint by explicitly regularizing the $l_k$ norms between the reconstructed and real images. However, as the data distribution becomes more complicated like in CIFAR-10, the $l_k$ norms become inadequate to measure the reconstruction. Consequently, ALICE's reconstruction error on CIFAR-10 increases significantly compared with that on MNIST. In contrast, the reconstruction performance of DALI is consistent on both data sets. The reason is that our model explicitly specifies the dependency structure of the generative model, and matches both prior and conditional distributions without using the simple data-fitting $l_k$ metrics in the data space. This can be further justified by the performance of DALI-$l_2$ which follows the same structure. Compared with DALI-$l_2$, DALI further decreases the MSE significantly by a relative 49\% on CIFAR-10, which shows that the inferred conditional variance is crucial for achieving the faithful reconstructions on complicated data sets.

\begin{figure*}[htb]
\centering
\begin{subfigure}{0.32\textwidth}
\includegraphics[width=1.67in]{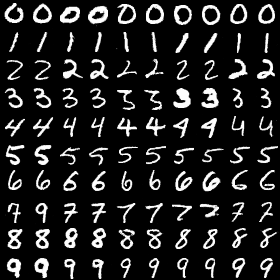}
\end{subfigure}
\begin{subfigure}{0.32\textwidth}
\includegraphics[width=1.7in]{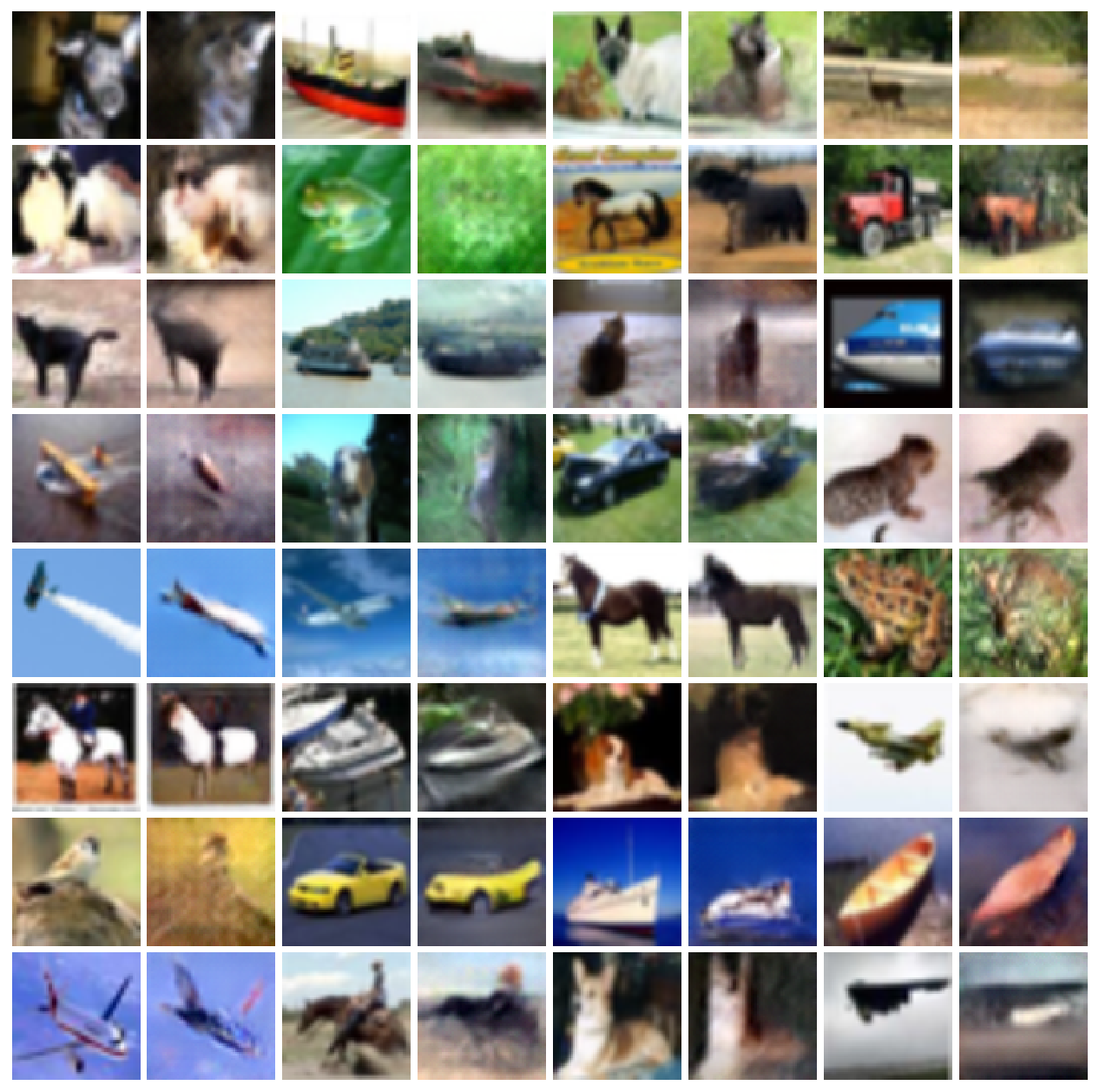}
\end{subfigure}
\begin{subfigure}{0.32\textwidth}
\includegraphics[width=1.7in]{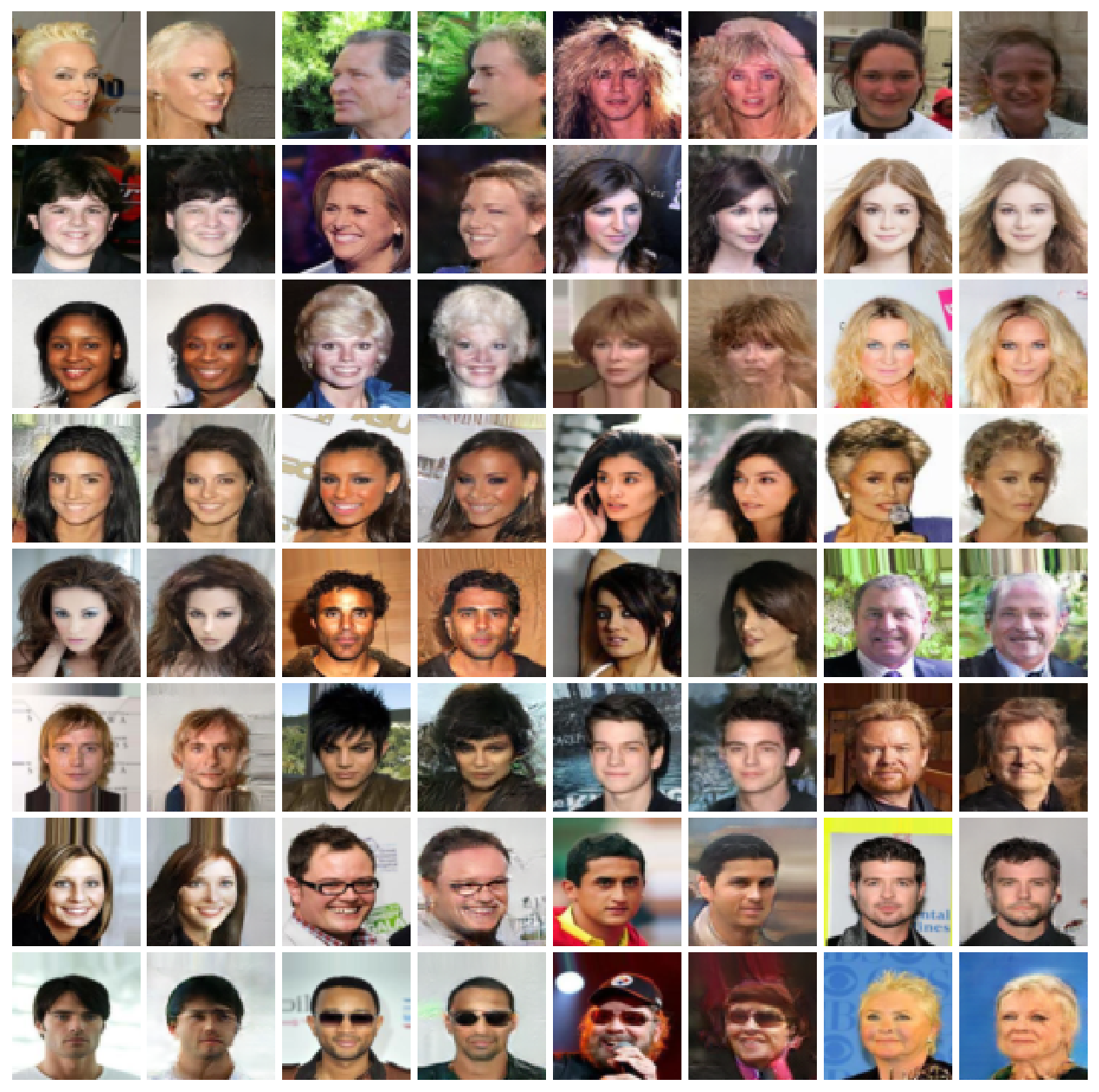}
\vspace{0.15in}
\end{subfigure}

\vspace{-0.0in}
\begin{subfigure}{0.32\textwidth}
\includegraphics[width=1.67in,height=1.65in]{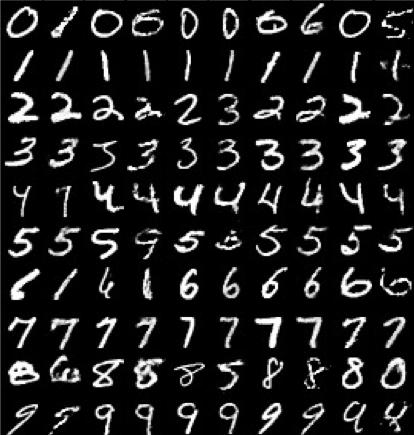}
\end{subfigure}
\begin{subfigure}{0.32\textwidth}
\includegraphics[width=1.7in]{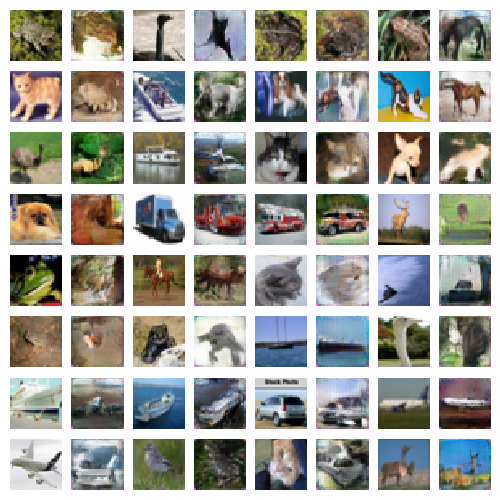}
\end{subfigure}
\begin{subfigure}{0.32\textwidth}
\includegraphics[width=1.7in]{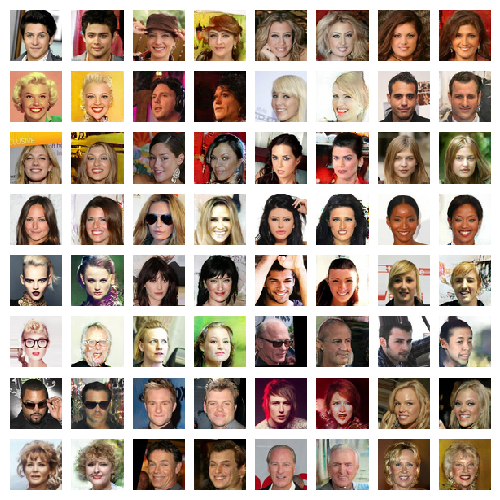}
\end{subfigure}
\caption{\label{fig:reconstruction}Reconstruction comparison between our proposed model DALI (first row) and ALI (BiGAN) (second row) on MNIST, CIFAR-10 and CelebA datasets. In each subfigure, the odd columns represent original samples from the test set and the even columns are their reconstructions.}
\vspace{-0.0in}
\end{figure*}

\paragraph{Generation} DALI outperforms all the baseline models including GAN on inception score. This suggests that DALI can bring further improvement on generation performance instead of deteriorating it like the other baselines. The reason that both ALI and ALICE perform worse than GAN on generation is that the task of matching two complicated joint distributions, $p_{\theta}(\bz, \bx)$ and $q_{\phi}(\bz, \bx)$, is more difficult than the task of the regular GAN, which is to match only the marginals, $p_{\theta}(\bx)$ and $q_{\phi}(\bx)$. The proposed model DALI explicitly defines the dependency structure between $\bz$ and $\bx$, which is more effective compared with one step joint distribution matching. Comparison between DALI and DALI-$l_2$ shows that the learned variance is also critical for better generation performance. 
We also want to highlight that DALI's generation performance can be further improved by replacing the adversarial network with more advanced state-of-the-art GANs. 

\subsection{Visualization of the Reconstructions}

In Figure \ref{fig:reconstruction}, we compare reconstruction of DALI with the results reported in ALI\cite{dumoulin2016adversarially} (BiGAN \cite{donahue2016adversarial}). 
From the first column of Figure \ref{fig:reconstruction}, we observe that ALI provides a certain level of reconstructions. However, it fails to capture the precise style of the original digits. In contrast, DALI can achieve very sharp and faithful reconstructions. 
On CIFAR-10, ALI's reconstructions are less faithful and oftentimes make mistakes in capturing exact object placement, color, style, and object identity. Our model produces better reconstructions in all these aspects. 
For the reconstructions on CelebA, DALI reproduces the similar style, color and face placement, and even achieves a high level of face identity. As stated in \citet{dumoulin2016adversarially}, they believe ALI's unfaithful reconstructions is caused by underfitting. This also leads us to believe that our adversarial regime (marginal and conditional distribution matching) is more efficient for inference compared to joint distribution matching regimes.

\subsection{Mode Collapse Reduction}

\begin{table*}[!htb]
\centering
\caption{Degree of mode collapse, measured by modes captured~(higher is better) and \% high quality samples~(higher is better) on 2D grid data. The baseline results of GAN, ALI and Unrolled GAN are reported in \protect\citet{srivastava2017veegan}. }
\label{tab:mode_collapse}
\begin{tabular}{lcccccccr}
\toprule

                         &GAN  & ALI   & Unrolled GAN & VAEGAN & VEEGAN & SN-GAN      & DALI          & DALI-$f$   \\
\midrule

Modes (Max 25)           & 3.3 & 15.84 & 23.6         & 21.4   &  24.6  & \textbf{25} & \textbf{25}   & \textbf{25} \\
\% High Quality Samples  & 0.5 & 1.6   & 16           & 34.1   &  40    & 67.8        & \textbf{81.1} & 66.4 \\

\bottomrule
\end{tabular}
\end{table*}




To show the effectiveness of our model on mode collapse reduction, we perform the same synthetic experiment as in \citet{dumoulin2016adversarially}. The data is a 2D Gaussian mixture of 25 components laid out on a grid. To quantify the degree of mode collapse, we use the two metrics used in \citet{srivastava2017veegan}: the \textit{number of modes captured} and the \textit{percentage of high quality samples}. A generated sample is counted as high quality if it is within three standard deviations of the nearest mode. Then the number of modes captured is the number of mixture components whose mean is nearest to at least one high quality sample. We compare the proposed method DALI and DALI-$f$ to ALI, Unrolled GAN \cite{metz2016unrolled}, VAEGAN \cite{larsen2015autoencoding}, VEEGAN \cite{srivastava2017veegan} and SN-GAN \cite{miyato2018spectral}. As shown in Table \ref{tab:mode_collapse}, the proposed model DALI provides the best performance on both measures consistently. More specifically, DALI can capture 25 modes every time and generate more than 80\% of high-quality samples. This suggests that the proposed model DALI significantly alleviates the mode collapse issue of the GAN framework and hence further improves the generation performance. 

\section{Related Work}
The most straightforward way to learn an inference mechanism is to learn the inverse mapping of GAN's generator post-hoc \cite{zhu2016generative}. However, since its training process is the same as GAN, it still suffers from mode collapse problem. InfoGAN \cite{chen2016infogan} minimizes the mutual information between a subset $\mathbf{c}$ of the latent code and the generated samples, and hence can only do partial inference on $\mathbf{c}$. AGE \cite{ulyanov2017takes} encourages encoder and generator to be reciprocal by simultaneously minimizing an $l_1$ reconstruction error in the data space and an $l_2$ error in the code space. This is closely related to the cycle-consistency criterion \cite{zhu2017unpaired,kim2017learning,yi2017dualgan,li2017alice}. Although the pairwise reconstruction errors help reduce mode collapse, the data reconstruction is still measured by $l_1$ or $l_2$ norm, which brings the same problem of VAE and VAE-GAN hybrids. It is worth noting that the main difference between our method and VAE is not about which divergence we use, but rather about upon which space we calculate the divergence. In VAE, they calculate the divergence on $\bz$-space, but in DALI, we calculate the divergence on $\bx$. Putting (reverse) KL-divergence on $\bx$ allows us to play the adversarial game on the more complicated distribution of $\bx$, but leave the parametric reconstruction to simpler $\bz$.

Different from the heuristic combination of VAE and GANs, \citet{mescheder2017adversarial} theoretically derived an adversarial game to replace the KL-divergence term in the variational lower bound (also called ELBO), and gives the new method, adversarial variational Bayes (AVB), much more flexibility in its dependence on latent $\bz$. However, the reconstruction term on $\bx$ still exists and so is the parametric assumption on the conditional data distribution, leading to the blurriness in their reconstructed and generated samples.

ALI \cite{dumoulin2016adversarially,donahue2016adversarial} is an elegant approach to bring inference mechanism into adversarial learning without assuming parametric distribution on the data. Different from our work, it directly plays an adversarial game to match the joint distributions of the decoder and encoder. But in practice, ALI's reconstructions are not necessarily faithful because the dependency structures within the two joint distributions are not specified \cite{li2017alice}. ALICE \cite{li2017alice} tries to solve this problem by regularizing ALI using an extra conditional entropy constraint on the data. The conditional entropy is either explicitly measured by $l_k$ norm, or implicitly learned by adversarial training. However, when the data distribution becomes complicated (e.g. CIFAR-10), the $l_k$ metric may lead to blurry reconstructions and the adversarial training is hard to achieve \cite{li2017alice}. Compared with ALI and ALICE, our method is proven to minimize the KL-divergence between both priors and conditionals of generator and encoder, and can provide consistent effective inference even on complicated distribution (see Section \ref{subsec:quant}).

\citet{srivastava2017veegan} proposed VEEGAN to tackle the mode collapse issue of GANs by adding an implicit variatinoal learning on the latent $\bz$. To our best knowledge, this is by far the only approach that is also reconstructing $\bz$. Different from VAEs, VEEGAN autoencodes the latent variable or noise $\bz$. By doing so, it enforces the generator not to collapse the mappings of $\bz$ to a single mode, because otherwise, the encoder will not be able to recover all the noise $\bz$. The details of their model can be summarized as ALI regularized by an extra reconstruction of latent $\bz$. Therefore, VEEGAN is similar to ALICE in the sense that they are both adversarial games on the joint distribution with an extra regularization on either data or latent reconstruction. Our model DALI instead only plays the adversarial game on the marginal data distribution, and reconstructs the latent $\bz$ by maximizing its log-likelihood under the latent posterior distribution.

\section{Conclusion and Future Work}
We proposed a novel framework, DALI, which matches both prior and conditional distributions between the generator and the encoder. Adversarial inference is incorporated into this framework and there is no parametric assumption on the conditional data distribution. We show in the experiments that the proposed method not only allows efficient inference but also improves the image generation. 

The assumption on $q_{\phi}(\bz|\bx)$ can be further released using an autoregressive $p_{\theta}(\bz)$. However, the same technique cannot be easily applied to $q_{\phi}(\bx)$ or $p_{\theta}(\bx|\bz)$. Therefore, we believe the reconstruction direction $\bz \to \bx \to \bz$ is more expressive than the opposite $\bx \to \bz \to \bx$.

\newpage
\small
\bibliographystyle{named}
\bibliography{ijcai20}

\end{document}